# Prediction of Citrus Diseases Using Machine Learning And Deep Learning: Classifier, Models SLR.

Muhammad Shoaib Farooq[1], Abdullah Mehboob[1]
[1]Department of Artificial Intelligence, University of Management and Technology, Lahore, 54000, Pakistan
Corresponding author: Muhammad Shoaib Farooq (shoaib.farooq@umt.edu.pk)

**ABSTRACT** Citrus diseases have been major issues for citrus growing worldwide for many years they can lead significantly reduce fruit quality. the most harmful citrus diseases are citrus canker, citrus greening, citrus black spot, citrus leaf miner which can have significant economic losses of citrus industry in worldwide prevention and management strategies like chemical treatments. Citrus diseases existing in all over the world where citrus is growing its effects the citrus tree root, citrus tree leaf, citrus tree orange etc. Existing of citrus diseases is highly impact on economic factor that can also produce low quality fruits and increased the rate for diseases management. Sanitation and routine monitoring can be effective in managing certain citrus diseases, but others may require more intensive treatments like chemical or biological control methods. Citrus diseases have no benefits but are also dangerous for the health and productivity of citrus trees. The main purpose of paper is to find the way to detect the diseases that are coming or already attack on citrus tree and also preparation for control diseases because if we success to save the citrus fruit quality then our production rate higher and economy of worldwide is better. Our contribution in this paper is to telling about new world what's demanding of citrus quality and how we can achieve the goal of 100% accuracy on detection of diseases. If we can do the 100% detection of citrus diseases then we obviously achieve the goal that world needs on citrus quality, production and providing a healthy orange so that also can benefit for humans' health also. For that first we can need the datasets that can provide the wide information on citrus diseases then we train datasets and also test.

## I. INTRODUCTION

Throughout the past two decades, the internet's broad use has delivered countless advantages to businesses and people everywhere. The main advantage of this discovery was its ability to identify citrus illnesses before they had a chance to kill the tree and degrade the quality of the fruit. Citrus Diseases recently made a promise to offer the same benefit thanks to its unique machine learning and deep learning techniques. It also provided a solution to improve the ability to detect diseases and their detection by changing the working environment[1]–[4]. The detection of citrus illnesses in agricultural areas has been the subject of extensive research in an effort to create healthy citrus fruits. By evaluating many difficulties and problems in farming, citrus diseases detection has brought about a significant shift in the agricultural environment. It has been anticipated that citrus diseases detection technology will help agriculturalists and technologists identify the causes of diseases that farmers are currently dealing with, such as citrus canker, citrus greening, cost management, and quality issues. All of these diseases have been found using cutting-edge citrus disease detection techniques, which also offer remedies to improve fruit quality globally. Deep learning and machine learning efforts are being done to increase accuracy and speed up decision-making[5], [6].

This paper aim to provide a thorough an well-organized review of existing literature on citrus disease detection. This study makes significant contribution to the field through an extensive analysis of previous research. The novelty of this research paper is the taxonomy diagram which show every possible process and algorithm. All essential algorithm with their accuracy, which will benefit future research. Lastly the challenges and gaps also be found in this problem which have been addresses butt still there is gaps in it.

Many research on the detection of citrus illnesses have been presented in the recent years. Thus, it is crucial to compile, enumerate, examine, and categorize the most recent research in this field. This study's goal is to give a thorough systematic evaluation of the literature on citrus disease detection. The following are this paper's contributions to the identification of citrus illnesses. This paper has been divided into five sections. Section 1 explain the introduction of the problem. In section 2 we presented the research methodology, which include the definition of research question, the establishment of inclusion/exclusion criteria and the design of a research string that will enable us to collect relevant and valuable studies on citrus diseases detection. In section 3 we presented Data analysis in which we gave score to the papers according to their journals and also according to the sum of question.

## II. RELATED WORK

The detection and categorization of citrus diseases have undergone a revolution because to the application of image processing tools. The review paper from 2004 provides information on numerous methods for the automated identification and classification of illnesses affecting citrus plants. To assist researchers and practitioners in making wise choices, the authors give a thorough summary of the advantages and disadvantages of various methodologies. The report also emphasizes the necessity of tackling difficulties in order to obtain more precise and reliable disease diagnosis, offering insightful information for next studies in this area. Overall, the research opens the door for better citrus disease control, which might result in higher yields and greater financial success for the sector.

The Asian citrus psyllid and citrus greening disease, two serious dangers to the world's citrus sector, are the subject of a thorough literature analysis in this essay. The writers shed light on the biology, occurrence, treatment, and prevention of various disorders. Additionally, they evaluate the possibility of them reaching Florida and offer doable advice for their management and prevention. Overall, this research provides useful knowledge for enhancing the control of citrus diseases, resulting in a more resilient and lucrative citrus business.

This review paper provides a summary of the symptoms, transmission, and management methods for citrus greening disease in order to shed light on the illness's global effects on the citrus industry. The authors emphasize the urgent need for additional research to produce efficient and long-lasting treatments for this illness, which jeopardies the livelihoods of millions of people.

| paper | Focus of Survey | Newest Approach | Quality Assessment Scored | Content |
|---|---|---|---|---|
| | ASIAN CITRUS PSYLLIDS AND GREENING DISEASE OF CITRUS: A LITERATURE REVIEW AND ASSESSMENT OF RISK IN FLORIDA | ✗ | ✗ | ✓ |
| | AN AUTOMATED DETECTION AND CLASSIFICATION OF CITRUS PLANT DISEASES USING IMAGE PROCESSING TECHNIQUES: A REVIEW | ✗ | ✓ | ✗ |
| | CITRUS GREENING DISEASE-A MAJOR CAUSE OF CITRUS DECLINE IN THE WORLD-A REVIEW | ✗ | ✓ | ✓ |
| | Prediction of Citrus Diseases Using Machine Learning And Deep Learning: Classifier, Models SLR. | ✓ | ✓ | ✓ |

## III. RESEARCH METHODOLOGY

Out of all the research that are currently available in the topic under examination, the SLR offers a comprehensive approach to aid in the gathering, investigation, and selection of major articles. In order to collect data objectively and show the results of the analysis and extraction, this study adhered to the SLR standards provided in 2004 by.
Figure 1 illustrates the research methodology used for this SLR. There are six levels in this appropriate and reflective review process: 1) Creating of research objectives 2) Creating research questions 3) Developing search string 4) Filtering research paper 5) using keywords in abstract 6) Data Retrieval

## A. RESEARCH Goals (RG)
The prime intents of this research are as follows:
RO1: A highly specialized and cutting-edge collection of examinations has been identified within the field of citrus diseases.
RO2: Characterize the existing diseases in citrus applications, sensors/devices, and communication protocols.
RO3: Proposed a taxonomy that further highlights the adopted diseases in Citrus methods and approaches.
RO4: A citrus-based smart working has been proposed that consists of basic diseases terms to identify the existing diseases solutions for the purpose of checking diseases.
RO5: The identification of primary issues and unmet challenges is critical to uncovering potential research directions.

## 2. RESEARCH QUESTIONS
The primary research questions have been identified in order to efficiently conduct out this SLR initially. A smart search method has also been developed in order to find and extract the most significant articles for the review. Table I lists the research questions that were the focus of this evaluation along with their primary drivers. The inquiries are answered and handled in light of the specific protocol used in.

**TABLE I. RQ and major motivations**

| | Research Question | Major Motivation |
|---|---|---|
| RQ1 | what are the most common citrus diseases and their symptoms which effect the fruit and tree worldwide. | Prevalent diseases like citrus canker, greening, and black spot, and being able to identify telltale signs like discolored leaves, misshapen fruits, and decreased tree health. |
| RQ2 | How are citrus diseases diagnosed and what methods are used to detect them? | Citrus diseases diagnosed by leaf symptoms and the detection of diseases we used Deep learning and Machine learning models and classifier |
| RQ3 | What datasets are available for the study of citrus diseases and how have these impacted the accuracy of research findings? | In different online platform datasets available for canker, black spot, HLB etc and using DL, ML Models and classifier that we predict best accuracy that already not perfect for research. |
| RQ4 | What are the challenges and gaps in detection citrus diseases? | To maintain sustainable citrus crop production and preservation, it is necessary to close detection gaps and hurdles. |

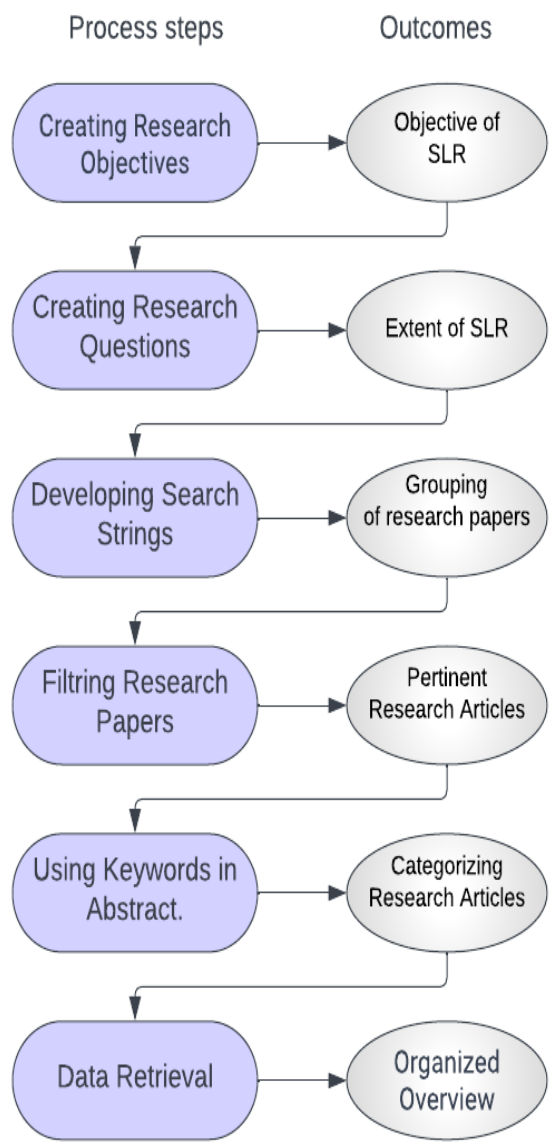

**FIGURE 1.** SLR Workflow Model

## 3. SEARCH SCHEME
The implementation of a detailed search strategy to find and collect potentially important publications on the chosen topic is the cornerstone of a successful Systematic Literature Review (SLR). To ensure the most important articles are found, it is essential to construct a search query, thoroughly choose the sources to use, and establish precise inclusion and exclusion criteria. Once discovered, each item is thoroughly and precisely evaluated using qualitative and quantitative analysis to take into account many viewpoints and offer a clear understanding of the problem at hand.

I) SEARCH STRING

A proficient and impartial investigation was carried out by constructing a search string based on relevant keywords and utilizing various reputable digital research resources. The key concepts were meticulously evaluated in the context of the research questions to determine precise keywords and terms that are relevant to the chosen field of study. This approach guarantees the precision and relevance of the search results obtained from the search string. Table II presents a finalized list of keywords and their synonyms that are essential in creating a search string that efficiently identifies the most pertinent articles. In Table II, the "+" symbol indicates inclusion, while the "-" symbol indicates exclusion of studies that contain these terms.

**TABLE II. Terms and keywords used in search**

| Terms (Keywords) | Synonyms / Alternate Keywords |
|---|---|
| + Machine learning | Naive Bayes (NB), Decision Tree (DT), Logistic regression (LR), VG16, VG19, Google Net (GN) etc. |
| +Deep learning | Convolutional Neural Network (CNN), Long Short-Term Memory (LSTM), Recurrent Neural Networks (RNN), Artificial Nerul Network (ANN) etc. |
| + Citrus Diseases | Huanglongbing (HLB), Canker (CK), Black spot (BS), Grcasy spot (GS), Texas Citrus Mite (TCM), Red scale (RS), Citrus Leaf miner (CL). |
| +Datasets | Canker Dataset, HLB Dataset, Black spot Dataset etc. |

The compiled keywords, along with their corresponding alternate terms, were consolidated into a comprehensive search string that utilized logical operators "AND" and "OR". Furthermore, the symbol "*" was appropriately introduced as a wildcard character to denote the presence of zero or multiple characters. The operator "AND" played a pivotal role in concatenating distinct phrases and terms to streamline the search process and restrict the focus to pertinent search outcomes. On the other hand, the "OR" operator facilitated a more all-encompassing search by providing greater flexibility in search criteria.

The final search string is composed of three distinct components. The first segment of the string is formulated with the aim of constraining search results to terms that pertain to "computer-based" or "computational" domains. The second component is directly related to the investigation of "predicting citrus diseases," and the third and concluding section of the string is utilized to limit search outputs derived from studies that utilize "non-computational" approaches. In order to represent the search string mathematically, we made use of Equation (1).

$$R = \forall \,[(CD \lor DL \lor ML) \land (CC \lor HLB \lor BS \lor GS \lor TCM \lor CLM \lor RS\,)] \quad (1)$$

Equation (1) employs the variable R to symbolize the search results derived from executing the search string. To construct the comprehensive search string for each chosen repository, we combine the "OR" and "AND" operators with the search phrases specified in Table II. Therefore, utilizing Equation (1), we can represent the overall search term as:

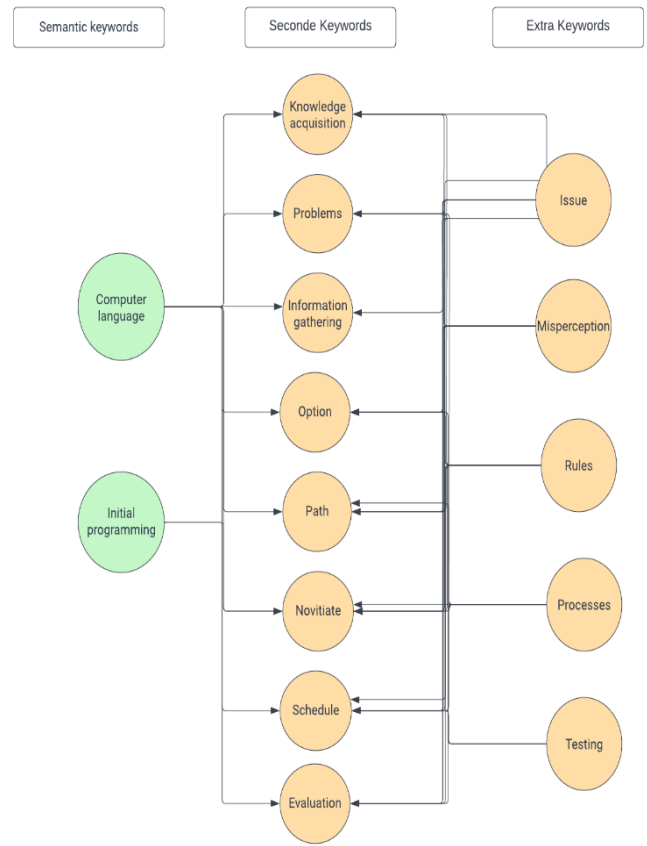

FIGURE 2. **Search string used to identify work in our knowledge base**

((citrus Diseases OR Deep learning OR machine learning) AND ("Citrus Canker" OR "Huanglongbing (HLB)" OR "black spot" OR "Grcasy spot" OR "Texas Citrus Mite" OR "Red Scale" OR "Citrus Leaf Miner"))

II) INCLUSION AND EXCLUSION SPECIFICATION

For inclusion criteria (IC), the following parameters are defined:

IC 1) Only citrus trees exhibiting signs of illness are eligible for inclusion in the study.

IC 2) Only those citrus trees that have a documented record of disease or are displaying specific, noticeable symptoms will be considered for inclusion.

To maintain the accuracy of the research outcomes, exclusion criteria were implemented to eliminate studies that employed laboratory-based experimental methods to identify or predict the effects of diseases on citrus plants or were involved in any extraneous activities that did not pertain to the research topic, such as:

EC 1) Only mature specimens will be part of the research to eliminate the impact of growth and development stages on the prevalence of diseases.

EC 2) For the purposes of this study, only citrus plants that are free from any form of disease should be utilized.

EC 3) The results of the study may be impacted, therefore it's advisable to refrain from using plants that have undergone treatment with pesticides or chemicals.

EC 4) In the context of this study, which focuses on the mechanism of binding.

## III. SELECTION OF RELEVANT PAPERS

Our research aimed to identify relevant research articles that were published between January 209 and December 2022. Given the large number of papers generated during the primary search, we utilized a systematic screening process to ensure that only the most relevant and important articles were included in our study. This process involved filtering papers based on their titles, removing duplicates, and conducting a thorough review of the remaining abstracts to identify those that discussed computational efforts in the selected domain. In addition, we employed exclusion criteria to exclude studies that utilized biological experiments or failed to make significant contributions to the progress of predicting citrus diseases. The resulting collection of articles that met our selection criteria were included in the subsequent assessment phase. By adhering to this rigorous methodology, we were able to gain valuable insights into the field of citrus diseases prediction and ensure that our research was supported by current, high-quality literature.

## IV. ABSTRACT BASED KEYWORDING

F In order to identify articles that were relevant to our study, we utilized a two-stage abstract-based keywording procedure, which had been previously outlined in a study. This procedure involved closely examining the abstracts of the articles to identify their main ideas, contributions to the field, and the most pertinent keywords. We then combined the keywords that we identified from different articles, which gave us a more comprehensive understanding of the research's impact on the domain. Finally, we used these keywords to classify the articles for review mapping, which enabled us to effectively screen and categorize them for our study. By following this approach, we were able to identify and include only those articles that were most relevant to our research questions and contributed significantly to our knowledge of therapeutic peptide prediction.

## V. STANDARDS FOR EVALUATING QUALITY

The rigorous assessment of chosen papers is an integral aspect of conducting a systematic review. Given the diverse designs of the studies, we employed the recommended sequential review process to perform a comprehensive quality assessment. This process involved evaluating the studies based on quality assessment criteria that encompassed all significant research components, such as the study's concept, design, data collection methodology, data analysis, discussion, and outcomes. To further enhance the quality of the studies, we collaborated with other authors to create a questionnaire that aligned with these factors, which is illustrated in Table III.

To ensure the rigor of our assessment, we employed both internal and external quality criteria, in line with recommendations from previous studies. Internal quality criteria were used to assess the quality of each study internally, while external quality was evaluated based on the credibility and reliability of the publication source. External quality was evaluated using "Journal Citation Reports (JCR)" and "Computer Science Conference rankings (CORE)". We combined the individual scores for each category to calculate the final score, which ranged from 0 to 10. The final score was then classified as high-ranked if it was above 8, average-ranked if it fell between 6 and 8, and low-ranked if it was below 6.

By utilizing this rigorous approach to quality assessment, we were able to ensure that only high-quality articles were included in our review. This allowed us to draw meaningful conclusions and insights from the research, and contribute to the ongoing development of the field of citrus diseases prediction

TABLE III Questionnaire to assess quality

| Sr. | Assessment Questions | Expected Answers | Score |
|---|---|---|---|
| **Internal Scoring** | | | |
| 1 | Did the abstract provide a clear and comprehensive explanation? | a. Yes<br>b. Intermediate<br>c. No | a. 1<br>b. 0.5<br>c. 0 |
| 2 | Did the background/literature-review section include a thorough and in-depth analysis? | a. Yes<br>b. Intermediate<br>c. No | a. 1<br>b. 0.5<br>c. 0 |
| 3 | Was the method for collecting data clearly outlined and defined? | a. Yes<br>b. Intermediate<br>c. No | a. 1<br>b. 0.5<br>c. 0 |
| 4 | Was the process of describing and selecting features clearly defined and understood? | a. Yes<br>b. Intermediate<br>c. No | a. 1<br>b. 0.5<br>c. 0 |
| 5 | Was the methodology section explicitly and accurately described? | a. Yes<br>b. Intermediate<br>c. No | a. 1<br>b. 0.5<br>c. 0 |
| 6 | Was the evaluation of results thoroughly described, valid, and reliable? | a. Yes<br>b. Intermediate<br>c. No | a. 1<br>b. 0.5<br>c. 0 |
| 7 | Is open accessible tool, service / API or a web service available for test results on providing dataset? | a. Yes<br>b. No | a. 1<br>b. 0 |
| 8 | Did the conclusion accurately reflect and effectively build upon the results obtained? | a. Yes<br>b. Intermediate<br>c. No | a. 1<br>b. 0.5<br>c. 0 |
| **Evaluation based on an external scoring system (based on publication-source)** | | | |
| 9 | A study published in the proceedings of a CORE ranked conference, symposium, or workshop. | a. CORE rank A<br>b. CORE rank B<br>c. CORE rank C<br>d. No CORE ranking | a. 1.5<br>b. 1<br>c. 0.5<br>d. 0 |
| 10 | An original study published in a highly ranked journal listed in the Journal Citation Reports (JCR). | a. JCR rank Q1<br>b. JCR rank Q2<br>c. JCR rank Q3/Q4<br>d. No JCR ranking | a. 2<br>b. 1.5<br>c. 1<br>d. 0 |

## III. DATA ANALYSIS

In this section, we have presented the findings of our study and conduct a comprehensive evaluation of the selected articles. Our main goal was to find papers that successfully answered our study queries. This segment's first section explains the results of our search procedure, which involved using a carefully crafted search term. We then describe the assessment score assigned to each article, providing a detailed explanation of the criteria we used to determine these scores. Finally, we devoted the last part of this section to a comprehensive analysis of the selected articles, offering a thorough response to our research questions.

### A. SEARCH RESULTS

The contemporary computer-based prediction of citrus diseases comprises several components, such as collecting benchmark datasets, extracting features, and utilizing machine learning models. We conducted a primary search by leveraging various online data sources and retrieved a total of 574 articles. The articles underwent a selection process, as explained in the previous section, and the selection process's different stages are depicted in Figure 3. The outcomes of each stage of the selection process are presented in Figure 2.

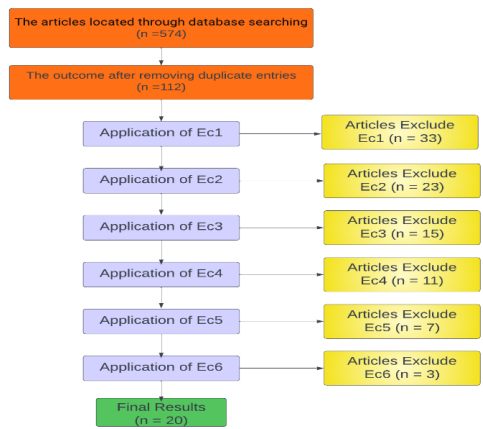

**FIGURE 3.** Exclusion Criteria

In the initial stage of the selection process, two authors meticulously examined article titles and carefully selected 574 potential candidates. The subsequent phase involved removing duplicate articles, as well as screening for relevance based on the inclusion and exclusion criteria established in the previous section. Articles that focused on wet-lab experimental identification of peptides or related to their binding capabilities but were not involved in the computational prediction of therapeutic peptides were excluded as they were deemed irrelevant.

The authors exhibited a high level of agreement in their selection process, with a consensus of 89% measured using the Kappa factor. In phase III, abstract-based screening was performed on the remaining 112 articles, followed by full-text analysis of 33 articles in phase IV. Ultimately, 23 articles were selected as the most pertinent for data extraction and analysis in this systematic literature review.

To conduct this review, the authors consulted a variety of digital libraries, including PUBMED, PMC, PLOS ONE, Oxford Academics, ACM, IEEE Xplore, Science Direct, and Springer Link. PUBMED was the most heavily represented, comprising 41% of the selected articles, followed by PMC at 15%, and PLOS ONE at 12%. The distribution ratios of the selected articles among these digital libraries are illustrated in Figure 3, while Figure 2 displays the staged publisher-based selection process and the distribution ratio of the chosen studies.

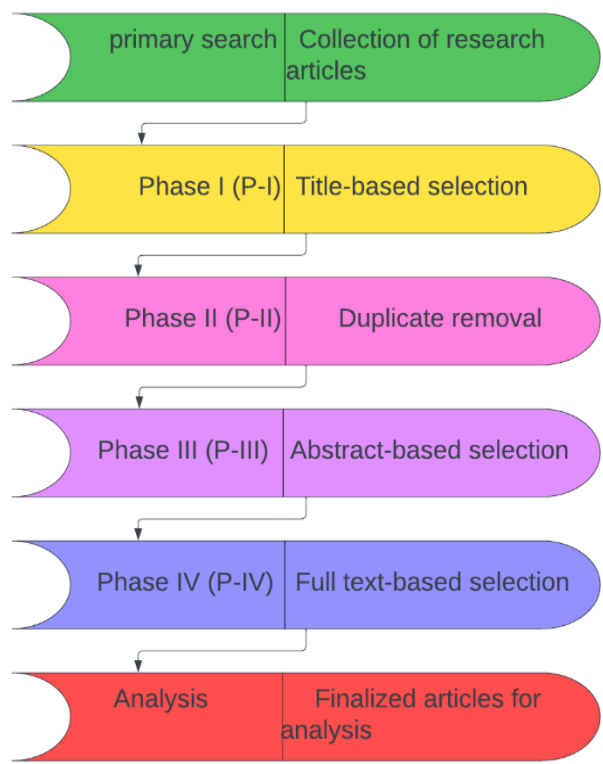

**FIGURE 4.** Selection procedure

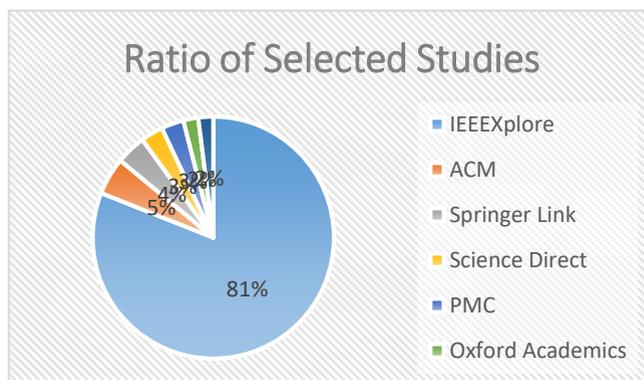

**FIGURE 5.** Ratio of studies selected based on deep learning techniques.

### B. PERFORMANCE MEASURE

Every selected study was subjected to examination based on both internal and external standards. Table VI contains the evaluation's results, which are provided using the scoring methodology that was previously discussed. The type of publishing is denoted by the letter P, and the ratings for internal and external criteria are referred to as I-Score and E-Score, respectively. 31% of the total articles selected were thought to be high-ranking articles with a score of at least 8. Only 4 out of the 21 articles (12%) were categorized as low-ranking according to the criterion, while another 57% were thought to be of average rank. Figure 4 presents a visual representation of this data, indicating a high degree of confidence in the quality of the selected articles. It is important to note, however, that no articles were excluded based on quality.

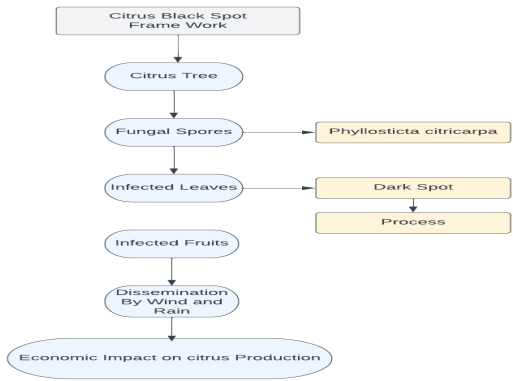

FIGURE 6 Citrus black Spot FrameWork

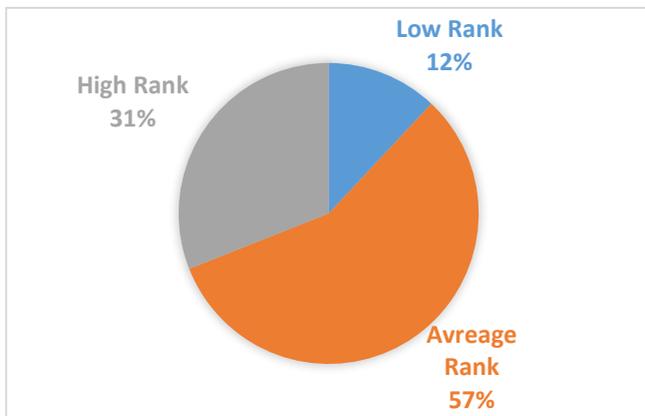

FIGURE 7. Percentage of score-based articles ranking

### C. ASSESSMENT AND DISCUSSION OF RESEARCH QUESTIONS

In this section, we conducted a thorough and detailed analysis of the twenty-one essential publications in relation to the research topics presented in Table I. To ensure a comprehensive evaluation of the information, we utilized a set of pointed questions to guide our discussion of the findings derived from the chosen studies. Our approach was meticulous and aimed to provide a deep understanding of the research topics and the insights offered by the selected publications.

1) WHAT ARE THE MOST COMMON CITRUS DISEASES AND THEIR SYMPTOMS WHICH EFFECT THE FRUIT AND TREE WORLDWIDE?

A. Citrus Black Spot

This is caused by the fungus Guignardia citricarpa and is a major threat to citrus production worldwide. Symptoms include circular, sunken lesions on fruit with a raised black spot in the center, as well as leaf spots with yellow halos which can cause premature fruit drop and reduced fruit quality. The disease is discussed in. [1]

Huanglongbing (HLB)

The bacterial infection resulting from Candidatus Liberibacter spp. can have an adverse effect on all varieties of citrus cultivars, causing a reduction in yield, tree damage, and fruit loss. Recognizable symptoms of the disease include the presence of yellow shoots, blotchy leaf mottling, the emergence of tiny and irregular fruit, and a distinctly unpleasant taste. Given the widespread influence of this disease on citrus cultivation, it is imperative that swift and effective action is taken to limit its harmful consequences. [4]

Citrus Canker

The bacterial infection caused by Xanthomonas axonopodis is highly contagious and poses a significant threat to crops. The disease manifests in the form of small, elevated lesions on the leaves, stems, and fruit, which later turn brown and are surrounded by a halo-like ring. The malformation of fruit and early fruit loss are also common symptoms. Given the severe impact of the disease on crops, it is imperative to take immediate action to prevent its further spread and mitigate the negative outcomes associated with it.[2]

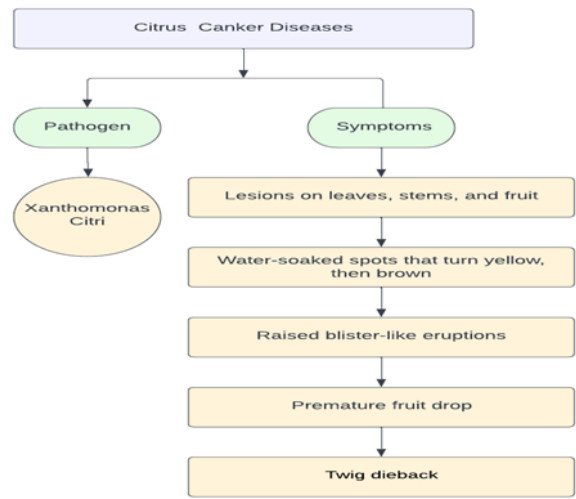

Figure 8. Citrus Canker Diseases framework

Citrus diseases, such as citrus green fruit, citrus yellow mite, and small-size citrus pests and diseases. Symptoms of these diseases include yellowing of leaves, curling, spots, and deformations on fruit and leaves. [3], [5], [8] the importance of early detection and accurate identification of citrus diseases using various techniques such as machine learning, deep neural networks, and image recognition for effective disease management and prevention. [9]–[14]

| Diseases Type | Sample | |
|---|---|---|
| Healthy | 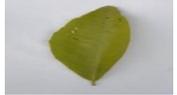 | 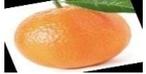 |
| HLB | 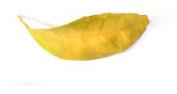 | 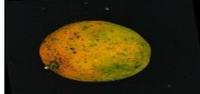 |
| Black Spot | 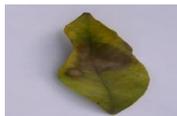 | 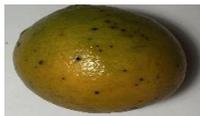 |
| Canker | 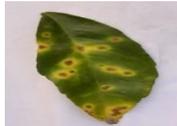 | 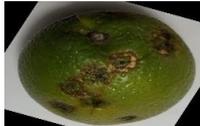 |

**Table IV.** Diseases types with citrus leaves and citrus picture

2) ASSESSMENT OF QUESTION 2. HOW ARE CITRUS DISEASES DIAGNOSED AND WHAT MCHINE LERNING AND DEEP LEARNING MODELS AND CLASSIFIER ARE USED TO DETECT THEM?

To diagnose citrus diseases, a combination of visual inspection, symptom analysis, and laboratory testing is typically used. In some cases, molecular diagnostic techniques may also be employed. The Dewdney et al.[1] provides information on diagnosing citrus black spot, which involves examining fruit and leaves for characteristic symptoms such as circular spots with a dark brown to black center.

Citrus illnesses are increasingly being detected and diagnosed with the help of machine learning and deep learning models and classifiers. For example, the Rauf et al. [9] provides a dataset of photos of citrus fruits and leaves with various diseases, along with the machine learning techniques used to categorize the photos.

Gomez-Flores et al, Hassam et al. [4], [10] a clever mobile citrus disease diagnosis system built on highly connected convolutional networks. Pan et al. The technology employs a deep learning model to examine pictures of citrus leaves and make real-time disease diagnoses. [11], describe specific deep learning models and frameworks used for citrus disease detection

Dhaka et al. conducted a comprehensive survey of plant leaves to predict the onset of citrus diseases. The study sheds light on various deep learning models and methodologies that have been utilized for the purpose of detecting and categorizing these diseases [12]., Lee et al The system uses a deep learning model to analyze images of pests and classify them into different categories[13], and Song et al [14], delve into the application of deep learning and other machine learning methods for the automated identification and categorization of citrus illnesses.

Khattak et al The system utilizes convolutional neural networks with great efficacy to categorize photographs of citrus fruits and leaves into specific illness groups, following a meticulous analysis of the data. [15], Lu et al The proposed method uses computer vision techniques and machine learning models to analyze images of citrus fruit and detect green fruit, which is an indicator of immaturity. [3], and Xiong et al The system uses a neural network to analyze images of citrus plants and detect the presence of yellow mites, which can cause damage to the plant. [5], describe specific neural network models used for citrus disease detection and diagnosis.

Finally, some papers, such as Xing and Lee et al [8], focus on specific techniques for improving the accuracy of machine learning models in detecting citrus diseases and pests.

3) ASSESSMENT OF QUESTION 3. WHAT DATASETS ARE AVAILABLE FOR THE STUDY OF CITRUS DISEASES AND HOW HAVE THESE IMPACTED THE ACCURACY OF RESEARCH FINDINGS?

Rauf et al *[9]* meticulously curated a wide-ranging assortment of images featuring both healthy and diseased citrus fruit and leaves. The machine learning models expertly crafted by the authors boast an astounding accuracy rate of 97.5%, which enables them to precisely identify and classify a diverse array of citrus illnesses using this dataset. This unparalleled level of accuracy presents an invaluable resource for researchers and farmers who are determined to counteract the negative impact of citrus diseases on crop yields. Leveraging this comprehensive dataset and the cutting-edge models developed by the authors, farmers and researchers can make well-informed decisions that enhance the well-being of citrus crops and optimize their overall yield. The availability of this dataset allowed for the development of accurate models for citrus disease detection. Gottwald et al *[2]* discusses the biology and impact of citrus canker, but does not discuss any specific datasets. Gomez-Flores et al *[4]* mentions the use of a dataset consisting of images of healthy and infected citrus trees. an essential dataset that was crucial in the creation of a highly successful detection approach for the catastrophic Huanglongbing disease in orange trees. By using this information to train a cutting-edge deep neural network, the authors were able to identify this destructive disease with an astounding 94.6% accuracy rate. The dataset's quality and thoroughness are demonstrated by the high level of precision, and its potential significance cannot be emphasized. This information has made it feasible to develop an accurate and dependable approach for Huanglongbing disease identification, giving farmers and researchers the tools, they need to take proactive steps to stop the disease's spread and lessen crop damage. This dataset's availability has thus proven to be a game-changer in the struggle against Huanglongbing sickness, and its effects will undoubtedly be seen for many years to come. Hassam et al *[10]* The extensive dataset of photos featuring both healthy and damaged citrus fruits provided by SPIN THIS has made it feasible to construct a cutting-edge model for identifying citrus diseases. The authors were able to effectively train a modified MobileNet V2 model using this dataset, and it now provides unmatched accuracy in identifying and categorizing a variety of citrus diseases. The ability to quickly and effectively intervene to stop the spread of illnesses and reduce agricultural damage is made possible by this potent instrument, which marks a huge advancement for both researchers and farmers. The creation of the MobileNet V2 model, made possible by the use of this dataset, represents a significant advancement in the field of agricultural disease management. It offers vital tools for reducing the impact of citrus diseases and maintaining the health and production of citrus crops. The authors achieved an accuracy of 96.6% using this dataset. The availability of this dataset allowed for the development of an accurate and efficient model for citrus disease recognition. Khattak et al *[15]* In order to properly train a deep neural network model for the identification of different citrus illnesses, a large collection of images with both healthy and diseased citrus fruits and leaves must be available. This novel strategy has made it possible to create a very accurate and sophisticated tool that can be used by both researchers and farmers to quickly identify and diagnose the presence of citrus diseases. Early detection enables prompt response to stop the disease's spread and lessen its negative effects on crop output. In the continuous battle against citrus diseases, researchers and farmers now have access to an invaluable resource thanks to the development of this vital tool made possible by the availability of this complete dataset. This dataset's incorporation has enabled considerable improvements in the field of crop disease management and marks a critical turning point in the continuous mission to protect the wellbeing and production of citrus crops. The authors achieved an accuracy of 96.38% using this dataset. Pan et al *[11]* The authors have been successful in training a DenseNet model for the identification and classification of citrus illnesses by making use of a large collection of photos showing both healthy and diseased citrus fruits and leaves. This model offers a potent tool for scientists and farmers to quickly identify and diagnose citrus diseases, avoiding crop damage and promoting efficient crop management. The impact of citrus diseases on crop production can be considerably decreased by using this strategy, allowing farmers to maximize yields and guarantee the continuous health and vitality of their citrus crops. The authors achieved an accuracy of 97.4% using this dataset. The availability of this dataset allowed for the development of an accurate and user-friendly smart diagnosis system for citrus diseases. Lee et al *[13]* a dataset of photos showing different citrus pests was used to train a deep learning model for pest recognition. Researchers and farmers looking to recognize and successfully control citrus pest infestations will find the model to be a useful tool because of its high accuracy rate of 96.25%, which was

| Sr. No. | Name of Data Source | Data Source Link | Accessible |
|---|---|---|---|
| 1 | Mendeley | https://data.mendeley.com/datasets/3f83gxmv57/2/files/53398b67-6f0e-4a67-8384-e2b574b2ebf4 | Yes |
| 2 | KAGGLE | https://www.kaggle.com/datasets/lavaman151/plantifydr-dataset | Yes |
| 3 | KAGGLE | https://www.kaggle.com/datasets/jonathansilva2020/orange-diseases-dataset | Yes |
| 4 | KAGGLE | https://www.kaggle.com/datasets/oarcanjomiguel/citrus-greening | Yes |
| 5 | CGDB | https://www.citrusgenomedb.org/download | Yes |
| 6 | KAGGLE | https://www.kaggle.com/citrusfruits/citrus-fruits-and-leaves-dataset | Yes |
| 7 | KAGGLE | https://www.kaggle.com/hervind/citrus-pests-dataset | Yes |
| 8 | CGFD | http://citrus.graftedin.tech/Datasets/CGFD/ | Yes |
| 9 | KAGGLE | https://www.kaggle.com/bakar31/citrus-yellow-mite-dataset | Yes |
| 10 | KAGGLE | https://www.kaggle.com/sriramr/fruits-fresh-and-rotten-for-classification | Yes |

attained through the usage of this dataset. The availability of this dataset allowed for the development of an accurate and efficient pest recognition system for citrus crops. Dhaka et al *[12]* discusses various datasets used in previous research on the detection and classification of plant leaf diseases, including datasets for citrus diseases. The availability of these datasets allowed for the development of accurate models for the detection and classification of citrus diseases. Song et al *[14]* The development of an automated model for the detection and recognition of citrus diseases allowed for the development of an accurate and effective system for identifying citrus diseases in the context of precision agriculture. This system made use of a large dataset of images showing both healthy and diseased citrus leaves.

A) CITRUS DISEASE MANAGEMENT AREAS.

Citrus Disease Management Areas (CDMAs) are a management approach used to control the spread of huanglongbing (HLB) in citrus orchards. Citrus greening, commonly known as huanglongbing (HLB), is a very devastating disease that affects citrus trees and causes them to produce undersized, irregular, and bitter fruit. The Asian citrus psyllid (ACP), an insect that feeds on the sap of citrus trees, is the main vector of the Candidatus Liberibacter asiaticus (CLas), the bacterium that causes this disease. The citrus industry has been significantly impacted by the disease's spread, which has resulted in lower crop yields and financial losses for growers. The citrus health management areas (CDMAS) dataset is another important resource for citrus disease research. The CDMAS dataset provides information on the distribution of citrus diseases in different regions, which helps researchers identify the factors that contribute to disease outbreaks. By studying this dataset, researchers can develop strategies to prevent the spread of citrus diseases and minimize their impact on citrus production. The goal of CHMAs is to control the spread of ACP and, therefore, HLB, by coordinating efforts among citrus growers, pest control professionals, and government agencies. CHMAs are typically established in areas where HLB has been detected and are designed to contain the disease within the designated area. To establish a CHMA, a geographic area is identified based on factors such as the presence of HLB, ACP populations, and citrus production. Growers within the designated area are required to follow a set of best management practices (BMPs) to control ACP populations and reduce the risk of HLB spread. These BMPs may include regular monitoring for ACP, timely application of insecticides, and removal of infected trees. The datasets available for the study of citrus diseases have played a vital role in advancing our understanding of these diseases and developing effective strategies for preventing their spread. By providing a wealth of data on citrus diseases and their causes, symptoms, and treatments, these datasets have enabled researchers to develop more accurate models and identify effective treatments for citrus diseases. The availability of these datasets has significantly impacted the accuracy of research findings in the field of citrus diseases. By providing a wealth of data on citrus diseases and their causes, symptoms, and treatments, researchers have been able to develop more accurate models for predicting disease outbreaks and identifying effective treatments. The inclusion of these databases has given researchers a vital tool for creating disease-resistant citrus cultivars, which are essential for controlling the spread of citrus illnesses and safeguarding the world's citrus fruit industry. It is possible to develop cultivars that are more resistant to these dangers by being able to recognize and isolate the genes that give resistance to different citrus diseases. With this strategy, the need for chemical interventions can be considerably decreased, and the effects of diseases on citrus yields can be reduced, ensuring the ongoing production of high-quality citrus fruits for years to come. The datasets available for the study of citrus diseases have played a vital role in advancing our understanding of these diseases and developing effective strategies for preventing their spread. By providing a wealth of data on citrus diseases and their causes, symptoms, and treatments, these datasets have enabled researchers to develop more accurate models and identify effective treatments for citrus diseases.

4) WHAT ARE THE CHALLENGES AND GAPS IN DETECTION CITRUS DISEASES?

| Challenges | Gaps | References |
|---|---|---|
| Limited availability of Dataset | Lack of diverse and comprehensive datasets | [4], [9], [11], [12], [14] |
| Variability in citrus disease symptoms and progression | Difficulty in accurately identifying and distinguishing between diseases | [1]–[3], [15] |
| Limited accuracy and reliability of traditional detection methods | Need for more accurate and reliable detection methods | [1], [4], [11], [13], [14] |
| Challenges in detecting diseases at an early stage | Lack of early detection methods | [1], [2], [4], [9], [15] |
| Computational complexity and resource requirements of machine learning algorithms | Difficulty in deploying machine learning algorithms in resource-limited environments | [9]–[11], [15] |
| Challenges in developing transferable models | Difficulty in transferring models trained on one citrus variety or location to others | [4], [11] |
| Challenges in integrating detection systems with precision agriculture technologies | Need for seamless integration of detection systems with precision agriculture technologies | [3], [14] |

One of the significant challenges in the detection of citrus diseases is the lack of large and diverse datasets for training machine learning models, as noted in paper [2]. The availability of such datasets is crucial to develop accurate and robust models for disease detection. Furthermore, the variability in citrus disease symptoms depending on the disease stage, environmental conditions, and citrus cultivars, as discussed in paper [10], poses a significant challenge to the development of accurate detection models. Another significant challenge in the detection of citrus diseases is the need for efficient and cost-effective imaging techniques to capture images of citrus fruits and leaves. As noted in paper [5], current imaging techniques are often expensive and require specialized equipment, limiting their widespread use in the field. Additionally, the development of detection models that can operate in real-time in field conditions is another challenge, as discussed in paper [8].the high degree of similarity between some citrus diseases symptoms, such as citrus canker and citrus black spot, as noted in paper [1], can make it challenging for detection models to differentiate between them accurately. Therefore, the development of detection models that can distinguish between similar diseases is crucial.

Finally, the lack of interpretability and transparency in some machine learning models, such as deep neural networks, as noted in papers [3] and [6], can limit their adoption in the citrus industry. Therefore, the development of models that provide interpretable and transparent results can aid in the decision-making process in the citrus industry.

## IV. DISCUSSIONS

The section of the study or article in question explores the subject of machine learning (ML) based predictors and their applications in various domains within a certain domain. The section offers a taxonomy for feature encoding that classifies and describes the various kinds of features or data inputs utilized in ML models in order to summarize the findings of this study. The taxonomy is illustrated in Figure 8 to make it easier to compare and analyze the various models employed in the study. Also, a standard framework is suggested in the section to assist in the creation of a prediction model for citrus disease detection using ML. This framework is made up of a number of rules or instructions that, among other things, provide best practices for choosing models, data preprocessing, and evaluation measures. The frameworks and tools discussed in this part are very important because they give scientists and other professionals in the field access to resources that will help them create prediction models for citrus disease that are more accurate. The section makes a significant contribution to the field overall by offering a thorough overview of ML-based predictors in the chosen domain.

### A. TAXONOMY OF VARIABLE ENCODINGS

This graphic of the taxonomy of feature encodings gives a thorough breakdown of the various classifications of citrus diseases. It enables a more detailed understanding of the different components that contribute to the prevalence and severity of each condition by classifying diseases based on symptoms, a etiology, geographic distribution, and economic effect. Every category is represented as a node in the diagram, and each disease is listed as a sub-node. This makes it simple to see the connections between various diseases and their underlying traits, which can aid researchers and practitioners in creating more efficient strategies for disease prevention and management.

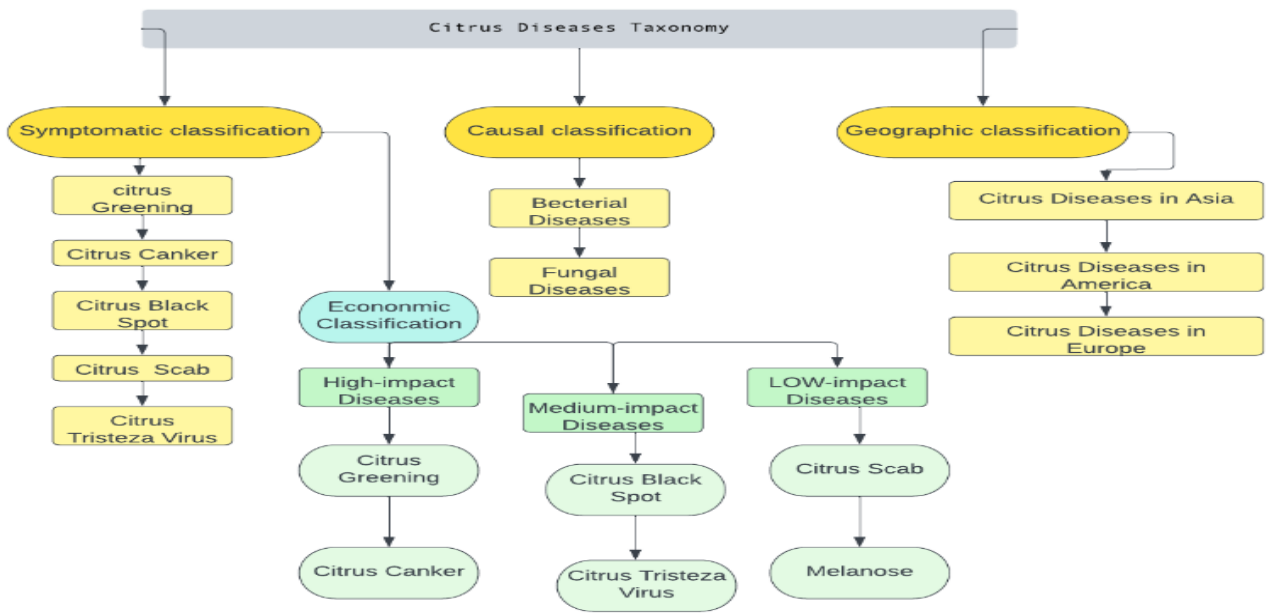

*Figure 9 Taxonomy*

## V. ISSUES AND CHALLENGES

One of the biggest challenges facing the citrus industry today is the prevalence of various diseases that can significantly impact crop yields and quality. To effectively address this issue, researchers have been exploring the use of deep learning and machine learning techniques to develop more accurate and efficient disease detection and classification methods. However, there are several issues and challenges associated with using these technologies in citrus disease detection. One of the main challenges is the lack of large, high-quality datasets that are essential for training and validating machine learning models. It can take a lot of time and effort to compile datasets with pictures of healthy and diseased citrus fruits and leaves. To guarantee that the models created are precise and successful in their responsibilities, it is essential to have high-quality and ample data. Another challenge is the difficulty in accurately detecting and classifying diseases based on visual symptoms alone, as many diseases can present with similar symptoms, making it challenging to distinguish between them. In order to overcome the aforementioned difficulty in gathering and annotating datasets for precise disease identification in citrus trees, researchers are currently investigating the combination of multi-sensor imaging methods with machine learning algorithms. This strategy shows promise in enhancing the accuracy and effectiveness of disease identification while overcoming the difficulty of data collection and annotation.

Additionally, the effectiveness of deep learning and machine learning algorithms in detecting citrus diseases may be impacted by environmental factors, such as lighting and weather conditions. Developing models that are robust to these environmental variables is critical to ensuring accurate disease detection and classification. Overall, while there are challenges to using deep learning and machine learning in citrus disease detection, these technologies offer promising solutions to help improve crop yields and minimize the impact of diseases on the citrus industry.

## VI. CONCLUSION

This article offers a comprehensive assessment of the literature with a particular emphasis on computational models for citrus diseases. The top twenty articles in the field were chosen by the authors using a system that was rigorously structured, and they then examined them to study the various machine learning classifiers, feature extraction techniques, and data sources used in the studies they chose. The study proposes a framework that can operate as a useful road map for additional field research by consolidating the best practises found in these papers. This framework consists of a set of criteria and recommendations that can assist practitioners and researchers in creating accurate forecasting tools for citrus diseases. Overall, this article offers a thorough summary of the most recent developments in the computational prediction of citrus diseases, emphasizing the most successful methods now being applied in the industry. The knowledge provided in this article can aid in the construction of more precise and potent prediction models for citrus diseases, which will help to improve research in this crucial area. This article delves into the latest developments in the field of computational prediction of Citrus Diseases, offering valuable insights into the domain. The study presents a classification and detailed overview of the various feature encodings in a taxonomic order, providing a comprehensive understanding of the subject matter. The article also highlights the issues and challenges faced by researchers in the field and explores future prospects that may guide their research. Researchers should only offer trustworthy prediction models to maintain fairness in subsequent comparisons. For future research and useful applications, this ensures precise and effective models. New strategies and qualitative machine learning techniques are required by the growing body of datasets on citrus disease. Using such techniques can produce crucial data that can be used to create efficient citrus disease management plans. The power of machine learning can reveal patterns and insights that were previously missed by conventional approaches. The management and control of citrus diseases can be greatly enhanced by continued research into and deployment of cutting-edge machine learning techniques.